\documentclass[letterpaper]{article} 
\usepackage{aaai25}  
\usepackage{times}  
\usepackage{helvet}  
\usepackage{courier}  
\usepackage[hyphens]{url}  
\usepackage{graphicx} 
\usepackage{natbib}  
\usepackage{caption} 
\usepackage{algorithm}
\usepackage{algorithmic}

\usepackage{multirow}
\usepackage{colortbl}
\usepackage[dvipsnames]{xcolor}
\usepackage{array}
\usepackage{soul}
\usepackage{xurl}
\usepackage{tikz}

\usepackage{amsfonts}

\usepackage{tabularx}
\usepackage{amsmath}
\usepackage{booktabs}

\usepackage{cleveref}
\newcommand{\thickhline}{\noalign{\hrule height.9pt}}

\usepackage{newfloat}
\usepackage{listings}
\DeclareCaptionStyle{ruled}{labelfont=normalfont,labelsep=colon,strut=off} 
\floatstyle{ruled}
\newfloat{listing}{tb}{lst}{}
\floatname{listing}{Listing}
\title{Foundation Models at Work: Fine-Tuning for Fairness\\in Algorithmic Hiring}
\author {
    Buse Sibel Korkmaz\textsuperscript{\rm 1\footnote{This work was done during an internship at IBM Research.}},
    Rahul Nair\textsuperscript{\rm 2},
    Elizabeth M. Daly\textsuperscript{\rm 2},
    Evangelos Anagnostopoulos\textsuperscript{\rm 3},
    Christos Varytimidis\textsuperscript{\rm 3},
    Antonio del Rio Chanona\textsuperscript{\rm 1}
}
\affiliations {
    \textsuperscript{\rm 1}Imperial College London\\
    \textsuperscript{\rm 2}IBM Research Europe\\
    \textsuperscript{\rm 3}Workable\\
    buse.korkmaz18@imperial.ac.uk, rahul.nair@ie.ibm.com, elizabeth.daly@ie.ibm.com, anagnostopoulos@workable.com, varytimidis@workable.com, a.del-rio-chanona@imperial.ac.uk
}

\begin{document}

\maketitle

\begin{abstract}

Foundation models require fine-tuning to ensure their generative outputs align with intended results for specific tasks. Automating this fine-tuning process is challenging, as it typically needs human feedback that can be expensive to acquire.
We present \emph{AutoRefine}, a method that leverages reinforcement learning for targeted fine-tuning, utilizing direct feedback from measurable performance improvements in specific downstream tasks. We demonstrate the method for a problem arising in algorithmic hiring platforms where linguistic biases influence a recommendation system. In this setting, a generative model seeks to rewrite given job specifications to receive more diverse candidate matches from a recommendation engine which matches jobs to candidates. Our model detects and regulates biases in job descriptions to meet diversity and fairness criteria. The experiments on a public hiring dataset and a real-world hiring platform showcase how large language models can assist in identifying and mitigation biases in the real world. We open-source our proposed method and related resources \footnote{\url{https://github.com/buseskorkmaz/FMs-at-work}}.

\end{abstract}

\section{Introduction}
\label{sec:intro}

Foundation models have demonstrated exceptional capabilities in generating coherent and contextually relevant text \citep{radford2019language, brown2020language, taylor2022galactica, thoppilan2022lamda, touvron2023llama, geminiteam2023gemini,jiang2024mixtral}. Large language models (LLMs) have accelerated progress in several natural language processing tasks, including text generation, translation, and sentiment analysis, among others \citep{liu2021examining, sallam2023chatgpt, lyu2023new}.

In practice, LLMs need to be adapted to specific tasks typically using a fine-tuning step. Fine-tuning aims to better align generative outputs on a specific task with desired outcomes. As a result, alignment research has emerged as a strategy that ensures the development of advanced AI systems resonates with intended goals and human values \citep{christiano2017deep, yuan2023rrhf}. 

One prominent approach here is Reinforcement Learning from Human Feedback (RLHF). By leveraging human demonstrations, preferences, or feedback, RLHF guides the fine-tuning process, allowing them to learn and approximate human values \citep{stiennon2020learning}. This method bridges the gap between human values and AI system behaviour, fostering a more robust and aligned decision-making process \citep{korbak2023pretraining}. However, RLHF is resource-intensive, requiring extensive human annotations.

In this paper, we present a strategy where fine-tuning (alignment) is driven by the downstream task directly, i.e. \emph{without human feedback}. We study this in the context of job description generation in hiring platforms where open jobs are matched to candidates using a recommendation system. We are interested in descriptions that appeal to a broad pool of candidates and do not marginalize specific groups.

Studies have shown that job postings using gender-neutral language have attracted a wider range of applicants than those with gender-biased terms \citep{woods2021chapter5}. Moreover, seemingly innocuous phrases in job descriptions can deter potential candidates, especially those from underrepresented groups \citep{woods2021chapter5}. For example, descriptions seeking ``young and energetic" candidates can dissuade older individuals and suggest a lack of flexibility for those with other commitments. 

In our setting, the risk of LLMs reinforcing societal stereotypes and prejudices is pronounced. LLMs can inadvertently inherit biases present in the underlying corpus \citep{bender2021dangers}. These biases can perpetuate unfairness, reinforce stereotypes, and marginalize certain social groups. For instance, language models trained on internet text data tend to exhibit gender and racial biases, leading to biased outputs when generating text or making predictions \citep{bolukbasi2016man, barikeri2021redditbias}. Recognizing and tackling these biases is essential to ensure fairness in various downstream tasks. 

Human preferences in this setting can be challenging to obtain from annotators for several reasons. Linguistic preferences are shaped by lived experiences that are varied \citep{davani2023disentangling}. There is an absence of normative descriptions that can be objectively judged. A job description that appeals to Alice may not appeal to Bob. Crucially, it is difficult for humans to reason about the likely impacts of their preferences when generative outputs are used in broader algorithmic settings. 

The objective of this work is to propose a framework for fine-tuning foundation models without human feedback, by quantifying their impact on downstream tasks. In our study, we examine the influence of linguistic biases on a recommendation system, taking job description generation as a concrete example. The diversity of candidates matched to these generated descriptions is a primary concern, emphasizing the need for fairness and bias mitigation. 

Our primary contributions are on: \emph{(a) task-responsive fine-tuning:} We propose a novel approach for fine-tuning foundation models using reward signals derived from measurable outcomes in downstream tasks. This method facilitates precise model adjustments based on actual task impacts, bypassing traditional reliance on human feedback collection or preference modelling, \emph{(b) bias mitigation for foundation models:} We demonstrate our method for bias mitigation and fairness when downstream tasks need to satisfy established equity criteria. As biases are quantified and mitigated during fine-tuning, our method ensures that generated content is purpose-aligned and inclusive, and \emph{(c) application to job description generation:} We present a real-world use case in job description generation, showcasing the practical challenges and our solutions to ensure that attract a diverse and fair candidate pool.

Related work is relegated to Appendix. We describe our method next.

\section{Method}

\begin{figure}[t]
    \centering
    \includegraphics[width=3.48in]{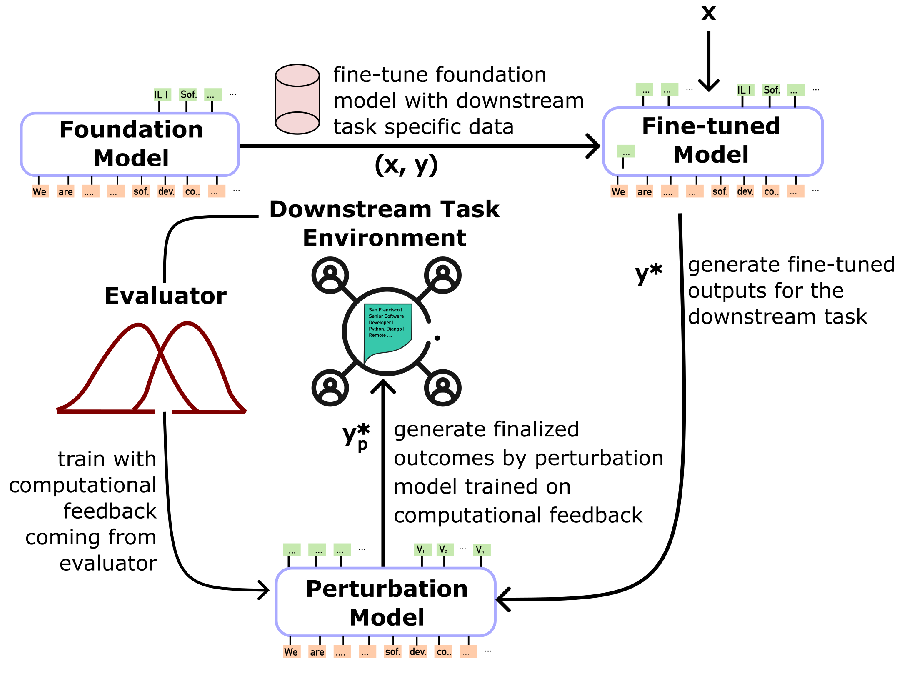}
    \caption{Our methodology \emph{AutoRefine} works by building a perturbation model that assesses the alignment of the generated content with task-specific goals. Evaluations serve as computational feedback that iteratively updates the perturbation model. During generation, both the original and perturbation models are used to generate tokens.} 
    \label{fig:high-level-diagram}
\end{figure}

We propose \emph{AutoRefine}, a method for fine-tuning foundation models without requiring human feedback. The process works in three stages: first aligning the model to a specific downstream task, then introducing a perturbation mechanism to optimize outputs based on performance metrics, and finally deploying the combined system. Below, we present our problem formulation and explain in detail how \emph{AutoRefine} reduces bias in job descriptions using feedback from a recommender system.

\subsection{Problem formulation}

Our framework consists of three core components: a pre-trained foundation model $\mathcal{M}$, a specific downstream task $ \mathcal{T}$, and a performance evaluator \( E \) that measures how well $\mathcal{M}$'s outputs perform. Our goal is to fine-tune $\mathcal{M}$ so its outputs for task $\mathcal{T}$ improve based on feedback from \( E \). We now describe each component and our training process in detail.

\subsubsection{Downstream task environment.} The downstream task environment represents the specific task \( \mathcal{T} \) for which the foundation model \( \mathcal{M} \) is being fine-tuned and optimized on. It takes the generated fine-tuned outputs \( y^* \) from the fine-tuned model \( \mathcal{M}^* \) as input and interacts with the evaluator \( E \) to assess the performance of the generated outputs.

The downstream task environment can vary depending on the specific application. For example, in our context of job description generation, the downstream task environment involves matching job descriptions with candidates. The task environment provides the necessary context and data for evaluating the generated outputs against the desired task-specific goals.

\subsubsection{Fine-tuned model.}

The first step in our training process is to fine-tune the foundation model $\mathcal{M}$ using a supervised dataset \( \mathcal{D} \) consisting of prompts \( x \) and their corresponding expected responses \( y \). This step is not fairness-aware and aims to fine-tune the model $\mathcal{M}$ to produce similar responses to $y$ for a given prompt $x$. The objective is to minimize:
\begin{equation}\label{objective}
\mathcal{L}(\mathcal{M}) = -\mathbb{E}_{(x,y) \sim \mathcal{D}}[\log P(y|\mathcal{M}(x))].
\end{equation}
Here, \( P(y|\mathcal{M}(x)) \) is the likelihood of \( \mathcal{M} \) producing \( y \) for prompt \( x \). We minimize \( \mathcal{L} \), ensuring the model's output aligns with the dataset. The fine-tuned model is defined as:
\begin{equation}\label{initial-fine-tuning}
\mathcal{M}^* = \arg\min_{\mathcal{M}} \mathcal{L}(\mathcal{M})
\end{equation}
where \( \mathcal{M}^* \) produces \( y^* \) for \( x \).

\subsubsection{Evaluator.}

The evaluator \( E \) serves as a computational feedback mechanism that assesses the performance of the generated outputs from the perturbation model \( \mathcal{M}_p^* \) in the context of the downstream task environment. It takes the generated outputs \( y^* \) from the downstream task environment \( \mathcal{T} \) and evaluates them based on predefined metrics. The evaluator provides a quantitative measure of how well the generated outputs align with the desired task-specific goals by assigning rewards or scores to the generated outputs, which serve as feedback signals to guide the training of the perturbation model.

In the context of fairness of job descriptions, the evaluator assesses the difference between the targeted and realized diversity metrics. The evaluator assigns higher rewards to job descriptions that meet the desired fairness criteria and lower rewards to those that lack diversity. The specific implementation of the evaluator may vary depending on the downstream task and the desired optimization objectives. 

The rewards or scores generated by the evaluator are then used to update the perturbation model \( \mathcal{M}_p \) through a reinforcement learning approach. The perturbation model learns to adjust the output probabilities of the fine-tuned model based on the feedback received from the evaluator. This iterative process allows the methodology to optimize the generated outputs towards the desired task-specific goals.

\subsubsection{Perturbation model.}

To train \( \mathcal{M}_p \), we employ a reinforcement learning (RL) approach. The perturbation model learns a value function \( V(y^*) \) that estimates the expected future reward for generating output \( y^* \). The value function is learned through interactions with the fine-tuned model \( \mathcal{M}^* \) and the downstream task evaluator \( \mathcal{T} \).

The perturbation model \( \mathcal{M}_p \) is defined as a function that modifies the output probabilities of \( \mathcal{M}^* \) based on the learned value function:
\begin{equation}\label{perturbation_model}
\mathcal{M}_p(y^* | x) \propto \mathcal{M}^*(y^* | x) f(\beta, V(y^*))
\end{equation}
where \( f(\beta, V(y^*)) \) is a function that takes the temperature parameter \( \beta \) and the value function \( V(y^*) \) as inputs and returns a non-negative value that scales the probabilities of the fine-tuned model \( \mathcal{M}^* \). In our implementation, we choose the exponential function \( f(\beta, V(y^*)) = e^{\beta V(y^*)} \) for its desirable properties and simplicity.

The training objective for \( \mathcal{M}_p \) is to maximize the expected reward while maintaining proximity to the fine-tuned model's output distribution:
\begin{equation}
\begin{split}
\mathcal{L}(\mathcal{M}_p) = \mathbb{E}_{y^* \sim \mathcal{M}^{*}(x)}[E(\mathcal{T}(\mathcal{M}_p(y^*)))].
\end{split}
\end{equation}

The training process involves iteratively generating outputs from the fine-tuned model \( \mathcal{M}^* \), applying the perturbation model \( \mathcal{M}_p \) to obtain perturbed outputs, evaluating the perturbed outputs using \(E\), and updating the value function and perturbation model based on the received rewards. During inference, the optimized perturbation model \( \mathcal{M}_{p}^{*} \) regulates the outputs of the fine-tuned model \( \mathcal{M}^* \) to generate better-aligned outputs with the downstream task goals.

\subsection{Reinforcement learning for fine-tuning}

In \emph{AutoRefine}, we employ RL to align the fine-tuned foundation model for a downstream task, similar to previous approaches \cite{bai2022constitutional, llms_human_feedback, askell2021general}. Our choice is motivated by RL's inherent capability to handle dynamic feedback and optimize non-differentiable objectives, aligning well with the performance evaluator \( E \) in our problem formulation. RL's strength in managing the exploration-exploitation trade-off ensures that our model generates diverse yet contextually relevant outputs. Given the sequential nature of text generation, we can model the prompt as a state and each generated token as an action and quantify the reward as feedback from the evaluator \( E \) for choosing a particular token given the state, without the need for explicit human annotations. This feedback shapes the reward function within the RL step of \emph{AutoRefine}, and any RL algorithm adept at managing such non-differentiable feedback can be employed in this stage.

We use the Implicit-Language Q-Learning (ILQL) algorithm \cite{snell2022offline} to train the perturbation model. ILQL is an offline reinforcement learning algorithm tailored specifically for language models. The primary advantages of ILQL in our context are twofold. Firstly, ILQL learns from a token-level Q-function which allows us to \textit{identify bias at the token level}. This representation offers a tangible metric that can be strategically optimized to manage and mitigate bias. Secondly, as being based on offline learning, it leverages samples from an existing dataset and \textit{reduces the queries to a recommendation engine} during training. Quantifying the reward for the generated text through the recommendation engine is a bottleneck in our setting due to the large size of the candidate pool such as a million candidates in the real-hiring platform data. 

In this step of \emph{AutoRefine}, the agent, which in our context is the language model, interacts with an environment to produce sequences, such as job descriptions. Feedback, in the form of rewards, is provided based on the inherent bias of the sequences produced. Q-value formulates the anticipated cumulative reward for a specific sequence. By leveraging this Q-value, the agent is trained to produce sequences that strike a balance between \textit{high quality and minimal bias}, similar to loss function-based debiasing techniques \citep{barikeri2021redditbias}.

\section{Debiasing Job Descriptions}\label{subsec:debiasing_job_descriptions}

Our goal is to create job descriptions that attract a broader pool of qualified applicants. In practice, algorithmic hiring tools recommend candidates for job openings through recommendation engines. For our research, we develop a proxy recommendation engine to serve as our evaluator $E$, simulating how such systems would operate in real-world recruitment platforms. We use \emph{AutoRefine} to generate job descriptions that achieve both effectiveness and inclusivity, refining the foundation model $\mathcal{M}$ based on feedback from this performance evaluator $E$.

The process begins with supervised fine-tuning (Equation \ref{objective}), which teaches the foundation model to generate appropriate job descriptions while maintaining its ability to produce coherent, relevant content. We then enhance these outputs by introducing carefully calibrated perturbations through $\mathcal{M}^*$, with the specific aim of improving diversity outcomes. The objective function for $\mathcal{M}_p$, detailed in Equation \ref{perturbation_model}, guides the model to make adjustments that better align the generated content with our diversity criteria. When deployed, this integrated system produces job descriptions that are both relevant to the position and meet established diversity standards.

For implementation, we selected GPT-2 as our base model and fine-tuned it on job description data, enabling it to generate appropriate descriptions from detailed job specifications. To address potential bias in the generated content, we then apply the second phase of \emph{AutoRefine}: training a perturbation model that incorporates feedback from the downstream task using ILQL \citep{snell2022offline}. During actual use, the system takes an original job description as an input. The perturbation model then generates an improved, less biased version by re-ranking the fine-tuned model's token outputs based on maximum potential reward. The following sections detail our reward function and explain how we optimize the learning process to effectively regulate application bias.

\subsection{Downstream task environment: Job-candidate matching engine}
To evaluate our approach, we develop a proxy recommendation system that simulates how algorithmic hiring tools match candidates to job openings. While bias in algorithmic hiring can originate from multiple sources, we specifically focus on bias stemming from job posting language rather than potential biases within recommendation systems themselves. This focus is important because job posting language has broader implications beyond algorithmic systems - it shapes how candidates perceive both the role and the company \citep{woods2021chapter5}. Our proxy system simulates both human application decisions and algorithmic hiring recommendations.

Our evaluation process has two main phases. First, the perturbation model modifies the fine-tuned LLM's token logits to reduce bias in the job posting. Then, we evaluate the rewritten description using our recommendation engine. This evaluation begins by filtering candidates based on the position's hard requirements. We then use BERT \citep{devlin-etal-2019-bert} to generate embeddings for both the job descriptions and candidate profiles. By computing cosine similarity between these embeddings, we identify the top $k$ candidates most similar to the job description. These candidates are then analyzed by our fairness evaluator to compute a diversity score.
\subsection{Fairness evaluator: Reward function driven by diversity}

Our approach evaluates fairness by comparing candidate distributions across two dimensions: gender and geolocation. We analyze these by measuring the difference between two probability distributions - the realized distribution \( D_{\text{realized}} \) from our selected candidates and a target distribution \( D_{\text{target}} \). To quantify this difference, we employ the 1-Wasserstein distance, which provides values between 0 and 1. Here, 0 indicates perfectly matching distributions, while 1 represents complete divergence (where one distribution is concentrated at 0 and the other at 1).

For gender analysis, we compute the 1-Wasserstein distance between the actual gender distribution of candidates matched to a rewritten job description  $y_{p}^{*}$, denoted as $D_{\text{realized, gender}}(y_{p}^{*})$, and our target distribution \( D_{\text{target, gender}} \):
\begin{equation}
\Delta_{\text{gender}}(y_{p}^{*}) = W_1(D_{\text{realized, gender}}(y_{p}^{*}), D_{\text{target, gender}}).
\end{equation}
The geolocation attribute $\Delta_{\text{geolocation}}(y_{p}^{*})$ can be computed similarly to the 1-Wasserstein distance between the realized and the target distributions.

These measurements combine to create our diversity score, which serves as the reward in our reinforcement learning environment. The score captures the total distribution mismatch across both attributes:
\begin{equation}
\mathcal{R}(y_{p}^{*}) = \Delta_{\text{gender}}(y_{p}^{*}) + \Delta_{\text{geolocation}}(y_{p}^{*}).
\end{equation}
A smaller Wasserstein distance between the achieved and target distributions indicates a higher diversity score, implying that the job description is more aligned with our diversity goals. 

\subsection{Metrics}

\subsubsection{Diversity score.} To evaluate our model's effectiveness, we apply \emph{AutoRefine} to rewrite job descriptions, focusing on the roles shown in Figure \ref{fig:genderimpact}. We compare diversity scores between original and rewritten descriptions. The effectiveness of our bias regulation is demonstrated when rewritten descriptions show smaller gaps between observed and target distributions compared to the original descriptions.
\subsubsection{Impact ratio.} We evaluate fairness using metrics established by New York Local Law 144 \cite{nyc144}, which provides a framework for auditing bias in algorithmic recruitment systems. The law introduces two key metrics to ensure transparency and equity: the \emph{selection rate} (measuring a cohort's historical success in being selected) and the \emph{impact ratio} (comparing a group's selection rate to that of the best-performing group). While the law covers multiple demographic categories including gender, race, ethnicity, and their intersections, our analysis focuses specifically on gender and location bias.
We implement these metrics in our setting as follows: For each group $g$ within a category $\mathcal{G}$ and for each job opening, we calculate selection rate using: (i) Selected candidates: The number of candidates from group $g$ appearing in the top-10 recommendations, (ii) Candidate pool: All relevant candidates from group $g$, where relevance is determined by cosine similarity (taking the top 50 candidates).

The selection rate for each group is calculated as:
\begin{equation}\label{sr}
\text{SR}_g = \frac{\text{top-10 candidates from $g$}}{\text{\# of relevant candidates from } g}
\qquad \forall g\in \mathcal{G}.
\end{equation}
The impact ratio (IR) is measured as the impact ratio relative to the best-performing group. 
\begin{equation}\label{ir}
\text{IR}_g = \frac{\text{SR}_g}
    {\max_{g^\prime \in \mathcal{G}}\text{SR}_{g^\prime}}
    \qquad \forall g\in \mathcal{G}.
\end{equation}

The best-performing group has $\text{IR}_g=1$. Values close to $1$ indicate equity across groups for that categorization. Values further away from $1$ indicate potential bias.

\subsubsection{TPR-GAP (True Positive Rate GAP).}
We adopt the fairness metric of TPR-GAP introduced by \citet{de2019bias} to our context. TPR represents the fraction of relevant candidates from a specific group $g$ that are included in the top-$k$ recommendations. The relevancy of a recommended candidate is decided by whether the profession of the matched candidate is the same position in the job advertisements.

For each group $g$ in a category $\mathcal{G}$, we define the True Precision Rate (TPR) as:
\begin{equation}\label{tpr}
\text{TPR}_g = \frac{\text{relevant candidates from $g$ in top-$k$}}{\text{total relevant candidates from $g$}}
\qquad \forall g\in \mathcal{G}.
\end{equation}

The TPR-GAP measures the difference in TPR between the best-performing group and the worst-performing group within a category $\mathcal{G}$:
\begin{equation}\label{tpr-gap}
\text{TPR-GAP} = \max_{g \in \mathcal{G}}\text{TPR}_g - \min_{g \in \mathcal{G}}\text{TPR}_g.
\end{equation}

A smaller TPR-GAP indicates higher fairness across groups, as it suggests that the top-$k$ recommendations include a similar proportion of relevant candidates from each group. Conversely, a larger TPR-GAP indicates potential bias, as it implies that the job-candidate matching system favors certain groups over others in terms of including relevant candidates in the top-$k$ recommendations.

\section{Experiments}

\newcommand{\change}[2]{%
 \tikz[baseline=(X.base)]{
   \node[
     rectangle,
     inner sep=1pt,
     fill=#1!20,
     anchor=base
   ] (X) {\footnotesize\textcolor{black}{#2}};
 }%
}
\newcommand{\increase}[1]{\change{green!50!black}{$\uparrow$#1\%}}
\newcommand{\decrease}[1]{\change{red}{$\downarrow$#1\%}}

\begin{table*}[t]
\small
\centering
\begin{tabular}{l rrr rrrr}
\toprule
& \multicolumn{3}{c}{Fairness} & \multicolumn{4}{c}{Text Quality} \\
\cmidrule(lr){2-4} \cmidrule(lr){5-8}
Model & Diversity Score & IR$_{\text{female}}$ & IR$_{\text{male}}$ & Naturalness & Coherence & Groundedness & Understand. \\
\hline
Original & -23.48 & 0.84 & 0.76 & 0.57 & 1.0 & 1.0 & 0.64 \\
GPT-2-large & \increase{7.4} -21.75 & \increase{2.4} 0.86 & \increase{1.3} 0.77 & \decrease{18} 0.47 & \decrease{3} 0.97 & \decrease{3} 0.97 & \decrease{17} 0.53 \\
\hspace{0.1em}+INLP-race & \increase{7.0} -21.83 & 0.84 & \increase{1.3} 0.77 & \decrease{70} 0.17 & \decrease{26} 0.74 & \decrease{70} 0.30 & \decrease{70} 0.19 \\
\hspace{0.1em}+INLP-gender & \increase{7.0} -21.83 & 0.84 & \increase{1.3} 0.77 & \decrease{70} 0.17 & \decrease{26} 0.74 & \decrease{69} 0.31 & \decrease{70} 0.19 \\
\hspace{0.1em}+SentD-race & \increase{7.0} -21.83 & 0.84 & \increase{1.3} 0.77 & \decrease{70} 0.17 & \decrease{26} 0.74 & \decrease{70} 0.30 & \decrease{70} 0.19 \\
\hspace{0.1em}+SentD-gender & \increase{7.2} -21.78 & 0.84 & \increase{2.6} 0.78 & \decrease{63} 0.21 & \decrease{78} 0.22 & \decrease{84} 0.16 & \decrease{64} 0.23 \\
\hspace{0.1em}+SD & \increase{6.6} -21.92 & 0.84 & \increase{2.6} 0.78 & \decrease{18} 0.47 & \decrease{3} 0.97 & \decrease{3} 0.97 & \decrease{17} 0.53 \\
\hspace{0.1em}+ID & \increase{6.2} -22.02 & \increase{2.4} 0.86 & \decrease{2.6} 0.74 & \decrease{37} 0.36 & \decrease{13} 0.87 & \decrease{16} 0.84 & \decrease{39} 0.39 \\
\hspace{0.1em}+\textbf{AutoRefine} & \increase{14.1} \textbf{-20.17} & 0.84 & \increase{2.6} \textbf{0.78} & \textbf{0.57} & \textbf{1.0} & \textbf{1.0} & \textbf{0.64} \\
\thickhline
\end{tabular}
\caption{Benchmarking of debiasing approaches comparing fairness and text quality metrics. Changes shown as percentages relative to original baseline. The highlighted (bold) scores show our method maintains original text quality while achieving the best improvement in diversity metrics.}
\label{table:combined-benchmarking}
\end{table*}

We conduct experiments on three datasets: the open-source dataset of Hackernews hiring posts and candidate profiles\footnote{\url{https://huggingface.co/datasets/dansbecker/hackernews_hiring_posts}}, Bias in Bios candidate profiles\footnote{\url{https://huggingface.co/datasets/LabHC/bias_in_bios}}, and a large dataset of job specifications and candidates from a real-world hiring platform\footnote{\url{https://www.workable.com/}}. The datasets are described in Appendix.

\subsection{Baselines}

We compare the performance of our proposed approach, \emph{AutoRefine}, with several debiasing algorithms. These algorithms represent different approaches to debiasing, ranging from embedding-level modifications to prompt-based techniques that we briefly introduce next. \citet{ravfogel-etal-2020-null} proposed Iterative Null-space Projection (INLP), a method for debiasing embeddings by iteratively projecting them onto the null-space of protected attributes. This approach aims to remove information related to sensitive attributes from the embeddings while preserving their utility for downstream tasks. \citet{liang-etal-2020-towards} introduced Sentence-Level Debiasing (SentD in Table \ref{table:combined-benchmarking}), an algorithm designed to debias pre-trained contextual embeddings at the sentence level, focusing on removing biases present in the representations of sentences and enabling more equitable downstream applications. \citet{schick2021self} developed Self-Debiasing (SD), a debiasing technique for GPT-2 models where the fine-tuned model self-diagnoses the bias present in the generated text and removes it, resulting in more neutral and unbiased outputs. Finally, \citet{morabito-etal-2023-debiasing} proposed Instructive-Debiasing (ID), an algorithm that utilizes debiasing prompts containing specific information about the category of bias present in a given text. By providing explicit instructions, the model learns to generate text that is less biased with respect to the specified categories. 

\subsection{Results}

\subsubsection{Fairness (Hacker News).} To assess fairness, we compare the diversity scores and impact ratios between the original and rewritten job descriptions for all tested methods.
Table \ref{table:combined-benchmarking} shows a 14\% improvement in the \emph{AutoRefine} rewritten descriptions compared to the original Hacker News posts. The reduced magnitude of the diversity score indicates a closer alignment with the desired diversity targets. Furthermore, the gender-specific impact ratios provide a more granular view of the alignment. For instance, the IR values for both male and female demographics remain consistent between the original and rewritten descriptions with a slight increase in $\text{IR}_{\text{male}}$. While our method is superior to other debiasing algorithms in terms of diversity score, the IR values of all methods are significantly close to each other. We also assess the IR focusing on geolocation, and the results do not exhibit significant differences among the methods. Nonetheless, we report these scores in the appendix for completeness. 

Table \ref{tb:examples} showcases specific examples from the evaluation set, highlighting the modifications made by the RL agent. The edits, though seemingly minor, have substantive implications for gender inclusivity and overall alignment with diversity goals. For instance, phrases that might be perceived as gender-biased or non-inclusive are either replaced or refined to ensure neutrality and inclusivity. Terms like ``maniacally focused on''  are changed to ``dedicated to", and specific gendered or potentially exclusionary terms are redacted or replaced, ensuring the descriptions are more universally appealing. More examples are given in Table \ref{additional-examples}. A key observation from the modifications made by the RL agent and their subsequent influence on the impact ratio is the profound effect of subtle changes on the downstream task. These nuanced alterations, despite their seemingly minor nature, can have significant effects. This phenomenon further underscores the challenges faced by human evaluators during feedback collection. Such subtle changes are often not easy to catch by human evaluators, emphasizing the intricacies of the task at hand and the need to develop alternatives to human feedback for fine-tuning foundation models.

\subsubsection{Fairness (Bias in Bios).}

As seen in Table \ref{tab:biasinbios}, the rewritten ads with fairness considerations reduce the TPR-GAP for 4 out of 5 occupations (except for the accountant role), which shows an improvement over the original job descriptions. We calculated the GAP as $\text{TPR}_{\text{female},y}$ -  $\text{TPR}_{\text{male},y}$ where $y$ is all considered occupations.

\begin{table}[tbh]
\small
\centering
\begin{tabular}{lrrr}
\thickhline
 & Original & AutoRefine \\ \hline
$\text{IR}_\mathrm{male}$ & 0.89 $\pm$ 0.20 & 0.97 $\pm$ 0.10 \\
$\text{IR}_\mathrm{female}$ & 0.70 $\pm$ 0.26 & 0.55 $\pm$ 0.24 \\
$\text{TPR-GAP}_\mathrm{\text{software-eng}}$ & 0.024 & 0.005\\
$\text{TPR-GAP}_\mathrm{attorney}$ & 0.009 & -0.007 \\
$\text{TPR-GAP}_\mathrm{accountant}$ & -0.006 & 0.023 \\
$\text{TPR-GAP}_\mathrm{professor}$ & 0.042 & 0.035 \\
$\text{TPR-GAP}_\mathrm{journalist}$ & -0.038 & -0.012 \\ \hline
Diversity Sc. & -6.135 & -5.93 \\
\thickhline
\end{tabular}
\caption{Key fairness measures on job rewriting experiments with Bias in Bios candidates dataset.}
\label{tab:biasinbios}
\end{table}

\subsubsection{Fairness (Hiring Platform Data).} We fine-tune GPT-2 using our proposed algorithm using both location and gender as diversity measures to optimize. In evaluation, we consider the impact ratio statistic for various groups of interest. We omit intersectional analysis for brevity. Table \ref{tab:workable} shows the impact ratio improving for female candidates. Job specification changes, however, had no impact on location-specific diversity in this instance. 

\begin{table}[tbh]
\small
    \centering
    \begin{tabular}{lllr}
\thickhline
 & Original & AutoRefine & $p$-value \\ \hline
$\text{IR}$\textsubscript{female} & 0.618$\pm$0.37 & 0.668$\pm$0.38 & 0.069* \\
$\text{IR}$\textsubscript{male} & 0.634$\pm$0.38 & 0.587$\pm$0.38 & 0.936 \\
$\text{IR}$\textsubscript{unknown} & 0.621$\pm$0.35 & 0.607$\pm$0.35 & 0.750 \\ 
$\text{IR}$\textsubscript{africa} & 0.250$\pm$0.40 & 0.181$\pm$0.36 & 0.995 \\
$\text{IR}$\textsubscript{asia} & 0.438$\pm$0.40 & 0.462$\pm$0.39 & 0.212 \\
$\text{IR}$\textsubscript{eu} & 0.364$\pm$0.39 & 0.322$\pm$0.39 & 0.995 \\
$\text{IR}$\textsubscript{na} & 0.563$\pm$0.31 & 0.586$\pm$0.32 & 0.522 \\
$\text{IR}$\textsubscript{oceania} & 0.252$\pm$0.42 & 0.265$\pm$0.43 & 0.334 \\
$\text{IR}$\textsubscript{sa} & 0.063$\pm$0.24 & 0.030$\pm$0.16 & 0.990 \\ 
\hline
Diversity Sc. & -20.96$\pm$9.83 & -22.67$\pm$9.51 & 0.961 \\
\thickhline
\end{tabular}
    \caption{Key fairness measures (mean $\pm$ standard deviation) on job rewriting experiments on Hiring Platform data before and after re-writes. Higher values are better. $p$-values from a binomial test showing the impact of rewrites (* implies significance at 10\%).}
    \label{tab:workable}
\end{table}

\newcommand{\valuechange}[2]{%
 \tikz[baseline=(X.base)]{
   \node[
     rectangle,
     inner sep=1pt,
     fill=#1!20,
     anchor=base
   ] (X) {\footnotesize\textcolor{black}{#2}};
 }%
}
\newcommand{\valueup}[1]{\valuechange{green!50!black}{$\uparrow${#1}}}
\newcommand{\valuedown}[1]{\valuechange{red}{$\downarrow${#1}}}

\begin{table*}[t]
\small
\centering
\begin{tabular}{ll rrrrrr}
\toprule
Profession & Type & MRR@10 & NDCG@10 & MRR@25 & NDCG@25 & MRR@50 & NDCG@50 \\
\hline
\multirow{2}{*}{Accountant} 
& Original & 0.614 & 0.848 & 0.614 & 1.466 & 0.620 & 2.265 \\
& Generated & 0.614 & \valueup{.059} 0.907 & \valueup{.007} 0.621 & \valueup{.052} 1.518 & \valueup{.001} 0.621 & \valuedown{.020} 2.245 \\
\multirow{2}{*}{Attorney}
& Original & 0.709 & 1.212 & 0.709 & 2.148 & 0.709 & 3.334 \\
& Generated & \valueup{.028} 0.737 & \valueup{.057} 1.269 & \valueup{.028} 0.737 & \valueup{.037} 2.185 & \valueup{.028} 0.737 & \valuedown{.154} 3.180 \\
\multirow{2}{*}{Journalist}
& Original & 0.920 & 1.314 & 0.920 & 2.222 & 0.920 & 3.369 \\
& Generated & \valuedown{.153} 0.767 & \valuedown{.034} 1.280 & \valuedown{.153} 0.767 & \valuedown{.097} 2.125 & \valuedown{.153} 0.767 & \valuedown{.252} 3.117 \\
\multirow{2}{*}{Professor}
& Original & 0.673 & 0.975 & 0.673 & 1.999 & 0.673 & 3.362 \\
& Generated & \valueup{.030} 0.703 & \valuedown{.054} 0.921 & \valueup{.030} 0.703 & \valuedown{.134} 1.865 & \valueup{.030} 0.703 & \valuedown{.135} 3.227 \\
\multirow{2}{*}{Software Eng.}
& Original & 0.086 & 0.140 & 0.092 & 0.275 & 0.097 & 0.472 \\
& Generated & \valuedown{.020} 0.066 & \valuedown{.039} 0.101 & \valuedown{.017} 0.075 & \valuedown{.070} 0.205 & \valuedown{.018} 0.079 & \valuedown{.112} 0.360 \\
\thickhline
\end{tabular}
\caption{Impact of fairness on recommendation quality across different professions and top-k values. Changes shown as absolute differences between original and generated versions.}
\label{tab:fairness_impact_full}
\end{table*}

In Table \ref{table:combined-benchmarking}, we present the significant advantage of our algorithm. Since our training includes supervised fine-tuning specific to the domain and then cleansing generated text with more application-oriented bias, \emph{AutoRefine} produces much less non-sensible response as a rewritten job description. To demonstrate that, we benefit UniEval \citep{zhong2022towards} language quality platform and evaluate language quality metrics in the aspects of naturalness, coherence, groundedness, and understandability for rewritten job descriptions of each algorithm in our benchmarking suite. Where it is required, we include the original job ad as a reference text in the language quality evaluation.

Our benchmarking results 
show that the proposed approach of implementing minimal token-based changes does not hurt the language quality while fairness metrics in Table \ref{table:combined-benchmarking} demonstrate our algorithm is competitive from the debiasing perspective. Moreover, the substantially worse language quality of the existing debiasing algorithms on text quality evaluation suggests the importance of reporting text quality-based scores for debiasing methods which alter language generation mechanics of underlying pre-trained methods and may degrade the quality of outputs.

\subsection{Impact of fairness on recommendation quality}

We investigate the potential impact of rewritten job advertisements on the utility of matching with the best and most qualified candidates. We evaluate the quality of the recommendations using metrics commonly employed in the recommender systems literature. Specifically, we consider Mean Reciprocal Rank (MRR) and Normalized Discounted Cumulative Gain (NDCG) to assess the effectiveness of matching job descriptions with candidate profiles.

Table \ref{tab:fairness_impact_full} presents the comparison of these metrics for the original job descriptions and our generated job descriptions across different professions at various top-k values (10, 25, and 50). The top-k values represent the number of top-ranked candidates considered for each job description. The results indicate that there is no substantial impact on the utility of the job descriptions after the rewriting process. In most cases, our generated job descriptions exhibit slightly higher match qualities compared to the original descriptions, as evident from the marginally higher values of MRR and NDCG across different top-k values.

For instance, considering the profession of accountant at top-10, the MRR@10 values for the original and generated descriptions are 0.614 and 0.614, respectively, and the NDCG@10 values are 0.907 and 0.848. These results suggest that the rewritten job descriptions maintain, and in some cases slightly improve, the quality of the recommendations.

However, it is worth noting that for the profession of software engineer, there is a slight decrease in the match qualities for the generated descriptions compared to the original ones. This can be attributed to the highly skewed distribution of the dataset for this profession, with an 84:16 ratio favoring males. Despite this, the overall impact on the recommendation quality remains minimal.

These findings demonstrate that our approach to promoting fairness in job advertisements does not compromise the utility of matching with the best and most qualified candidates. The rewritten job descriptions maintain comparable recommendation quality while addressing potential biases and promoting diversity in the candidate pool.

\section{Discussion}

Our work demonstrates a practical approach to implementing AI governance principles in the context of automated hiring systems, showcasing how technical solutions can support fairness and inclusivity while maintaining system effectiveness. This research bridges a critical gap between AI policy goals and technical implementation, particularly in the domain of algorithmic hiring where fairness considerations are of greatest importance.

The implications of our work extend in several important directions. First, it provides a concrete example of how large language models can be governed and aligned with societal values through automated feedback mechanisms, reducing reliance on human oversight while still maintaining accountability. Second, it demonstrates how technical solutions can help enforce policy objectives - in this case, fair hiring practices as outlined in regulations like New York Local Law 144. This alignment between technical implementation and policy requirements is crucial for effective AI governance.

Several limitations and considerations are important to note. While our methodology reduces the need for human intervention, it remains computationally expensive, highlighting the need to balance governance objectives with practical constraints. The choice of pre-trained model can significantly impact results, emphasizing the importance of model selection in governance frameworks. Additionally, our reliance on surrogate metrics rather than real-world outcomes points to the broader challenge of establishing appropriate evaluation criteria for AI governance mechanisms. The potential for factual errors in the generated descriptions also underscores the ongoing need for human oversight in AI systems.

\section{Acknowledgments}

This work was funded  by 
European Union’s Horizon Europe research and innovation programme 
under grant agreement no. 
101070568 (AutoFair).

\newpage

\bibliography{aaai25}
\newpage

\appendix
\section{Related Work}
\label{sec:related}

Our work relates to several lines of previous research. 

\subsubsection{Fair representations.} Early efforts in bias mitigation targeted removing gender biases in static embeddings where the semantic representation of a word is confined to just one vector like GloVe \citep{pennington-etal-2014-glove} and Word2Vec \citep{mikolov2013efficient} to achieve unbiased representations and word associations. \citet{bolukbasi2016man} studied how gender identity words are associated with specific occupations and subtracted the gender direction from word embeddings to neutralize the language while sustaining the equal distance between gender-neutral words and gendered pairs of words. \citet{ravfogel-etal-2020-null} proposed INLP for debiasing embeddings through iterative null-space projections for guarding protected attributes. As the field evolved, the focus shifted towards debiasing contextual embeddings, such as ELMo \citep{peters-etal-2018-deep}. \citet{kaneko2021debiasing, liang-etal-2020-towards} highlighted the relative complexity of contextual embeddings compared to their static counterparts and the challenge of identifying which parameters contribute to the bias. \citet{liang-etal-2020-towards} proposed SENT-DEBIAS applicable at sentence levels to debias pre-trained contextual embeddings, where \citet{kaneko2021debiasing} developed both token and sentence level approach and emphasized the trade-off between accuracy and unbiasedness in such models.

Several methods, such as adversarial learning \citep{elazar-goldberg-2018-adversarial, sakaguchi2021winogrande} and counterfactual data augmentation \citep{zmigrod2019counterfactual, islam2021fair}, have been proposed to mitigate language models propagating biases present in a training corpus. \citet{elazar-goldberg-2018-adversarial} investigated the encoding of demographic information in the intermediate representations learned by text-based neural classifiers. \citet{zmigrod2019counterfactual} aim to reduce bias by data augmentation through counterfactual statements. However, using augmented datasets is expensive and is prone to introduce noise and unrealistic scenarios that can negatively impact performance.

\subsubsection{Debiasing LLMs.}\citet{gira2022debiasing} proposed an LLM fine-tuning method by leveraging GPT-2 as an example, and showed their approach reduced the gender bias in GPT-2 on the StereoSet benchmark. While their method is relatively cost-effective compared to pre-training with an augmented dataset, the side effect of the fine-tuning for bias on downstream applications of language models has not been studied for their approach. \citet{schick2021self} developed a self-debiasing (SD) approach where a fine-tuned GPT-2 model self-diagnoses the bias and remove from the generated text. \citet{garimella2021he} suggested a combined method of further pre-training with fairness-aware datasets and then fine-tuning based on a loss function including regularizers for bias for BERT \citep{devlin-etal-2019-bert}. \citet{mao2023debiasing} noted the gap in existing works which separates the fine-tuning for debiasing then fine-tuning for downstream applications. Authors named the former bias as \textit{intrinsic bias} whereas the latter has been called as \textit{application bias} which our work targets to solve.   

\subsubsection{Measuring bias in LLMs.} A variety of benchmarks have been published to fairly evaluate and compare developed debiasing techniques aiming to address biases related to stereotypes or gendered word associations \citep{nadeem-etal-2021-stereoset, nangia-etal-2020-crows}. \citet{barikeri2021redditbias} targeted the bias in LLMs fine-tuned to conversational dialogue and have introduced RedditBias, a dataset rooted in actual Reddit conversations, designed to measure and mitigate biases in conversational models across gender, race, religion, and queerness. The study benchmarks the DialoGPT \citep{zhang-etal-2020-dialogpt} model with this dataset, revealing biases, particularly towards religious groups, and demonstrates that certain debiasing techniques can address these biases without sacrificing model performance.

\subsubsection{Prompt-based mitigation.} Recent studies investigated prompt-based fine-tuning strategies. \citet{guo2022auto} proposed extracting prompts from the pre-trained LLM \citep{devlin-etal-2019-bert, liu2019roberta},  through a beam-search style algorithm and then applying an equalizing loss over predicted token distributions. Their work demonstrated that the proposed bias mitigation strategy does not adversely impact the performance of LLM on downstream applications. \citet{morabito-etal-2023-debiasing} proposed an instructive-debiasing (ID) algorithm where debiasing prompts include specific information on the category of bias represented in the given text. \citet{si2023measuring} studied the effect of inductive biases through demonstrations without the model update in LLMs to overcome prior biases in an LLM. They conclude that intervention via inductive biases could be a helpful strategy to reduce the influence of feature biases, however, overcoming strong prior biases remains a challenging question on the topic of in-context learning.

\subsubsection{Fine-tuning.} Extensive research exists on the topic of model fine-tuning \citep{askell2021general, yuan2023rrhf}. \citet{llms_human_feedback} and \citet{liu2023summary} underscore the importance of RLHF in the fine-tuning process. While the models fine-tuned with RLHF have shown promise in generating more truthful and less toxic outputs, challenges persist. \citet{llms_human_feedback} emphasizes the balance between model alignment with human intent and maintaining high performance, highlighting the complexities and nuances of leveraging RLHF in the fine-tuning process. \citet{yuan2023rrhf} offers a critique on the RLHF approach, particularly highlighting the complexities associated with the PPO method \citep{schulman2017proximal}. \citet{casper2023open} survey the challenges and limitations of RLHF, specifically noting the difficulties attached to human evaluators, data quality and limitations of feedback types. Our approach addresses the concerns related to human evaluators as such eliminating the need for human feedback in fine-tuning. In some cases, \citep{bai2022constitutional, rafailov2023direct} LLMs are themselves used to produce alignment data used in fine-tuning. However, this can inadvertently reintroduce fairness and bias issues into the aligned LLMs sourcing from AI feedback.

From the hiring domain perspective, there is a large literature on fairness challenges in algorithmic hiring. We point to a recent survey \cite{fabris2023fairness}.

\section{Experiment Details}
\label{sec:experimentdetails}

\subsection{Datasets}
\label{app:subsec:datasets}
\begin{table}[ht]
    \centering
    \begin{tabular}{l|r|r} \thickhline
         Source & No. Jobs & No. profiles \\ \hline
         Hacker News & 76,000 & 20,300 \\
         Bias in Bios & 76,000 & 50,000 \\
         Hiring Platform & 5,745 & 1,000,000 \\ \thickhline
    \end{tabular}
    \caption{Summary of datasets used in experiments}
    \label{tab:datasets}
\end{table}

\subsubsection{Hacker News}
This dataset encapsulates a diverse collection of hiring posts from the tech news platform, Hacker News, with various splits, including ``hiring" (76K posts) and ``wants\_to\_be\_hired" (20.3K profiles). Each entry in the dataset contains a ``text" field, representing the content of the post, which corresponds to the job post in the ``hiring" split and candidate profiles in the ``wants\_to\_be\_hired" split. We curated a final version of the dataset for training, extracting specific useful features from the original job descriptions. The details of data processing and the implementation of ILQL, including hyperparameters, are detailed below.
We also conducted an analysis of the differential impact of gender identification on our embedding-based recommendation engine with Hacker News data. The results demonstrated that some roles are more susceptible to gender biases, details can be seen in  Figure \ref{fig:genderimpact}.

\subsubsection{Bias in Bios (Candidates)} 

We created a new dataset using a subset of the online biographies dataset introduced by \citet{de2019bias}. This dataset consists of over 400K biographies collected from the Common Crawl corpus, labelled with 28 different occupations. For our experiments, we focused on a subset of the test split, selecting biographies corresponding to occupations commonly found in job advertisements on Hacker News. This resulted in a candidate pool of 50K profiles. We then matched this candidate pool with both original and generated job descriptions from Hacker News and computed various diversity metrics, including the True Positive Rate Gender Gap (TPR-GAP). In this context, we consider true positive predictions as correct occupation matches between the labelled occupation of candidates and the occupations mentioned in the job advertisements, computed separately for each gender. 

\subsubsection{Hiring Platform Data.} 
The third dataset is from a live hiring platform. This dataset consists of 5745 job openings from around the world, with a very large pool of candidate profiles, obtained from publicly available data sources. 
Each job advertisement contains the job title, the description, and the requirements of the job. The jobs are distributed across different functions (e.g. engineering, accounting) and industries (e.g. software companies, food service). Each job is accompanied by a pool of 3K candidate profiles, both relevant and irrelevant to the job. The similarity of a candidate profile to the job is a float in $[0, 1]$ (higher values correspond to better matching candidates). For our experiments, we sample one million candidate profiles from this dataset randomly and use this subset as our common recommendation pool. 

\subsection{Processing Details}

\subsubsection{Hacker News Dataset}
We processed the original Hacker News dataset\footnote{\url{https://huggingface.co/datasets/dansbecker/hackernews_hiring_posts}} to make it more informative, structured and compatible with our environment. The details of processing for both job descriptions and candidate profiles are given next. 

\emph{Job descriptions:} The `text' column of the original `hiring' split includes job descriptions in an unstructured text format. We extract several features from this text to generate prompts to guide our approach to rewriting job descriptions while keeping the important job specifications in the rewritten text. We consider the job title, location of the opening, required technologies, the company offering the position, and if the remote working option is available as important features to use in the prompt. The template of `prompt' is as follows: 

\begin{quote}
    \textit{Original job description for reference: \texttt{$<$text$>$}. Based on this, the job is in \texttt{$<$location$>$}, at \texttt{$<$company$>$} for the \texttt{$<$job title$>$} position. The ideal candidate is skilled in \texttt{$<$technologies$>$}. \texttt{$<$Remote statement$>$} Write a new job description using only the original information.}
\end{quote}

Then, we calculated the diversity score of each job description using the recommender system, illustrated in Figure 2. 

\emph{Candidate profiles:} The `text' column of the original `wants\_to\_be\_hired' split includes candidate profiles in a relatively better-structured text format compared to job postings. While the dataset includes the location information of candidates, it doesn't contain any information related to gender which is an important feature to analyze the bias. Hence, we randomly assigned a gender for each candidate profile. The distribution of candidates based on geolocation and assigned genders is given in Table \ref{table:distribution}.

The embeddings of the both job descriptions and candidate profiles have been extracted from BERT \citep{devlin-etal-2019-bert} to use in recommendation engine. The splits with their engineered features are given in Table \ref{feature-extraction-table}.

\begin{table*}[tbhp]
\centering
\begin{tabular}{c|r|r|r|r|r|r|r|r|r|r}
\thickhline
 & \multicolumn{2}{c|}{Gender} & \multicolumn{8}{c}{Geolocation} \\ \cline{2-11} 
 & Female & Male & NA & Europe & SA & Asia & Africa & Remote & Australia & Unknown \\ \hline
Original & 0.5 & 0.5 & 0.55 & 0.21 & 0.03 & 0.1 & 0.01 & 0.01 & 0.01 & 0.06 \\ \thickhline
\end{tabular}
\caption{Distribution of candidate profiles over genders and geolocations. NA and SA represents north and south America, respectively.} 
\label{table:distribution}
\end{table*}

\begin{table*}[tbhp]
\centering
\begin{tabular}{l|l|l}
\thickhline 
Dataset & Feature & Extraction Method \\
\hline \hline
hiring & Job title & QA: What is the job title in the text? \\
\hline
hiring & Location & QA: What are the locations in the text? \\
\hline
hiring & Technologies & QA: What are the technologies in the text? \\
\hline
hiring & Company & QA: What is the company name in the text? \\
\hline
hiring & Remote & Text processing \\
\hline
hiring & Embedding & Extracted from BERT with given candidate profile \\
\hline
hiring & Prompt & Template: Job details (see caption) \\
\hline
hiring & Q-value & Diversity score obtained through recommendation engine\\
\hline
wants\_to\_be\_hired & Gender & Randomly assigned \\
\hline
wants\_to\_be\_hired & Location & Text processing \\
\hline
wants\_to\_be\_hired & Embedding & Extracted from BERT with given candidate profile \\
\hline
wants\_to\_be\_hired & Remote & Text processing \\
\thickhline
\end{tabular}
\caption[Caption for LOF]{Extracted features to obtain the final version of the dataset. QA represents the feature extraction methods using question-answering with Roberta\protect\footnotemark, where the context is always the original job description.}
\label{feature-extraction-table}
\end{table*}

\footnotetext{\url{https://huggingface.co/deepset/roberta-base-squad2}}

\subsubsection{Hiring Platform dataset}
\label{subsec:hiringplatform}

Data from the hiring platform consists of job profiles (Table \ref{app:workable:jobs}) and candidate profiles (Table \ref{app:workable:profiles}). The dataset consists of 5,745 job descriptions and several million candidate profiles from which we sample one million profiles for our experiments. The dataset is fully annoymised including the masking of institution names for experience and education history.

\begin{table*} [tbhp]
    \begin{tabularx}{\textwidth}{X|X|X|X}
    \thickhline
    \textbf{Variable Name} & \textbf{Description} & \textbf{Data Type} & \textbf{Example Values} \\ \hline
    id   &  Unique identifier for the job  & Integer  & 0 \\ \hline
    account\_id   &  Unique identifier of the account (company) that posted the job ad & Integer  & 23\\ \hline
    title   &  The title of the job  & str   &  Front-end Developer\\ \hline
    required\_experience   &  The level of experience required for the particular job  & str   & Mid-Senior level \\ \hline
    required\_education   &  The education level required for the particular job & str & Bachelor's Degree \\ \hline
    remote   &  Whether the job is remote or not & bool &  FALSE\\ \hline
    employment\_type  &  The employment type  & str & Full-time \\ \hline
    industry   &  The industry of the company that published the job ad & str  & Staffing and Recruiting \\ \hline
    function   &  The function of the particular job  & str  &  Engineering\\ \hline
    detailed\_location   &  The location of the company & JSON & \{"country\_code": [str] "IT",
"state\_code": [str] "MI", "city": [str] "Milan", "subregion": [str] “Metropolitan City of Milan”, "zip\_code": [str] “11111”\}
 \\ \hline
    description   &  The description of the job, containing details about the role & str & "We are looking for a ..." \\ \hline
    requirement\_summary   &  The requirements of the job & str & "Proven experience as ..." \\ \hline
    benefit\_summary   &  The benefits that the company will provide to the hired candidate(s) & str & "- Health Care Plan ..." \\ \thickhline    
    \end{tabularx}
    \caption{Hiring Platform Dataset - Job Advertisement. 
}
\label{app:workable:jobs}
\end{table*}

\begin{table*} [tbhp]
    \begin{tabularx}{\textwidth}{X|X|X|X}
    \thickhline
    \textbf{Variable Name} & \textbf{Description} & \textbf{Data Type} & \textbf{Example Values} \\ \hline
    id   &  Unique identifier for the profile  & Integer  & 0 \\ \hline
    job\_id   &  The ID of the job that the candidate corresponds to in the dataset & Integer  & 0\\ \hline
    country\_code   & The country code of the residence of the candidate & str   &  UK\\ \hline
    gender   & The predicted gender of the candidate & str   &  Male, Female, Unknown\\ \hline
    experiences   &  The work experiences of the candidate & JSON & \{"company": [int] 99, "start\_date": [str] "2020", "end\_date": [str] “2022”, "title": [str] "Graphic Designer"\} \\ \hline
    educations  &  The educations of the candidate & JSON & \{"school": [int] 10, "start\_date": [str] “2016”, "end\_date": [str] "2020", "field\_of\_study": [str] "Design",
"degree": [str] "Bsc"\},
 \\ \hline
    industry   &  The industry where the candidate has worked at & str  & Computer Software \\ \hline
    function   &  The function of the latest work experiences of the candidate & str  &  Engineering\\ \thickhline
    \end{tabularx}
    \caption{Hiring Platform Dataset - Profiles.}
    \label{app:workable:profiles}
\end{table*}

The gender of candidates is reported as male ($43.6$\%), female ($34.2$\%), or unknown ($22.2$\%). This forms the target gender distribution as this is considered to be the applicant pool.  Similarly, the location of candidates is available at the country level. This is aggregated by continent. The observed frequencies of candidates in this dataset serve as target location distribution. The reward model is additive in the Wasserstein distances for gender and location. 

The downstream task is one of matching candidates to job openings. For this we order candidates based on cosine similarity between the embedding of job description and candidate description. The candidate description is generated based on education and experience data on the candidate using templates. Both job and candidate embeddings are generated using a pre-trained model BERT.

The job descriptions are partitioned into train and $10$\% reserved for testing. The train data is further partitioned with $10$\% kept for validation. For our experiments, all one million candidates are considered to be viable for all job postings. In practice, this is not the case, as there may be filtering rules in place that limit the scope of the targeted audience. These constraints were not imposed.

For evaluation, we use the measures described in the main paper. Impact ratios, as codified in New York local laws, are retrospective, i.e. aim to audit hiring practices by examining hiring of candidates for each cohort. In our example, the hiring decision has not yet taken place. We consider a ``success'' if a candidate is ranked in the top ten for each job position. The denominator of the impact ratio, i.e. the applicant pool, is considered to be the top-50 applicants.
\subsection{Impact Ratios for Geolocation}\label{appendix:geolocationtable}

\begin{table*}[tbhp]
\centering
\begin{tabular}{l|r@{\hspace{0.75\tabcolsep}}r@{\hspace{0.75\tabcolsep}}r@{\hspace{0.75\tabcolsep}}r@{\hspace{0.75\tabcolsep}}r@{\hspace{0.75\tabcolsep}}r@{\hspace{0.75\tabcolsep}}}
\thickhline
 & \multicolumn{6}{c}{Geolocation} \\ \cline{2-7} 
 & IR\textsubscript{NA} & IR\textsubscript{Eu} & IR\textsubscript{Africa} & IR\textsubscript{Asia} & IR\textsubscript{SA} & IR\textsubscript{Remote} \\ \hline
Original &0.84$\pm$0.32&0.16$\pm$0.33&0.0$\pm$0.0& 0.11$\pm$0.30&0.05$\pm$0.21&0.01$\pm$0.11\\ \hline
GPT-2-large&0.84$\pm$0.31&0.15$\pm$0.33&0.0$\pm$0.0&0.1$\pm$0.3&0.07$\pm$0.24&0.02$\pm$0.11\\ 
\hspace{0.7em}+INLP-race&0.87$\pm$0.29&0.17$\pm$0.35&0.01$\pm$0.08&0.08$\pm$0.26&0.06$\pm$0.24&0.01$\pm$0.07 \\
\hspace{0.7em}+INLP-gender &0.87$\pm$0.29&0.17$\pm$0.35&0.01$\pm$0.08&0.08$\pm$0.26&0.06$\pm$0.24&0.01$\pm$0.07\\
\hspace{0.7em}+SentD-race &0.87$\pm$0.29&0.17$\pm$0.35&0.01$\pm$0.08&0.08$\pm$0.26&0.06$\pm$0.24&0.01$\pm$0.07 \\
\hspace{0.7em}+SentD-gender &0.86$\pm$0.3&0.17$\pm$0.35&0.01$\pm$0.08&0.08$\pm$0.26&0.06$\pm$0.24&0.01$\pm$0.1\\
\hspace{0.7em}+SD &0.83$\pm$0.32&0.18$\pm$0.36&0.0$\pm$0.05&0.11$\pm$0.3&0.05$\pm$0.22&0.01$\pm$0.11\\
\hspace{0.7em}+ID  &0.84$\pm$0.32&0.14$\pm$0.33&0.0$\pm$0.02&0.13$\pm$0.32&0.05$\pm$0.21&0.01$\pm$0.11 \\
\hspace{0.7em}+AutoRefine& 0.83$\pm$0.31&0.17$\pm$0.34&0.1$\pm$0.08& 0.12$\pm$0.32&0.02$\pm$0.12&0.03$\pm$0.16 \\ 
\thickhline
\end{tabular}
\caption{Geolocation-specific comparison of generated job description results. The reported score is the mean and standard deviation of the metrics over the evaluation dataset.} 
\label{table:geolocation-fairness}
\end{table*}

\subsection{Implementation}
\label{app:subsec:implementation}

We implemented the ILQL approach as detailed in \citep{snell2022offline}, using GPT-2 as our pre-trained LLM. Initially, GPT-2 was fine-tuned with original job descriptions to emulate the relationship between the provided prompt and generate detailed job descriptions. Subsequently, an RL agent was trained to determine the Q-value of the generated text and adjust subsequent token probabilities to meet diversity objectives. The `hiring dataset' was split into training, test, and evaluation subsets, while the `wants\_to\_be\_hired' dataset was used to compute diversity scores during both training and evaluation. To reduce the training time, 10\% of the hiring dataset has been used in experiments instead of entire dataset. 

For training, job descriptions were limited to 256 tokens, and the generated text was similarly restricted to 256 tokens. Within the recommender system, we set $k$ to $50$. During inference, we selected $\beta=64$, and results for varying $\beta$ values can be found in the next section. Our fine-tuned model was trained for 7 epochs with a learning rate of 1e-3, and the perturbation model was trained for an additional 7 epochs with the same learning rate. Altogether, the process took 10 hours using 8 V100 GPUs.

\subsection{Hyperparameter search}

We have reported the results and examples for $\beta=8$ in Results section. This value has been chosen intuitively with few trials in experiments. Here, we share the results for varying $\beta$ values in Table \ref{table:hyperparameter-search-2}. The results in Table \ref{table:hyperparameter-search-2} demonstrates that changes in $\beta$ do not cause significant deviations in the evaluations and our approach consistently outperforms the original dataset in terms of diversity score in all evaluated values of $\beta$ hyperparameter. 

\begin{table*}[tbhp]
\centering
\begin{tabular}{l|r@{\hspace{0.75\tabcolsep}}|r@{\hspace{0.75\tabcolsep}}|r@{\hspace{0.75\tabcolsep}}}
\thickhline
 & \multirow{2}{*}{\shortstack{Diversity \\ score}} & \multicolumn{2}{c}{Gender}  \\ \cline{3-4} 
 & & IR\textsubscript{female} & IR\textsubscript{male} \\ \hline
Original &-23.48$\pm$16.31&0.84$\pm$0.27&0.76$\pm$0.34\\ \hline
Rewrite ($\beta=2$) &-22.25$\pm$16.95&0.83$\pm$0.28&0.78$\pm$0.32 \\ \hline
Rewrite ($\beta=4$) &-22.17$\pm$17.00&0.83$\pm$0.23&0.78$\pm$0.32 \\ \hline
Rewrite ($\beta=8$) &-20.17$\pm$14.49&0.84$\pm$0.25&0.78$\pm$0.30 \\ \hline
Rewrite ($\beta=16$) &-22.05$\pm$17.02&0.84$\pm$0.26&0.78$\pm$0.32 \\ \hline
Rewrite ($\beta=32$) &-22.02$\pm$17.02&0.84$\pm$0.27&0.78$\pm$0.32 \\ \hline
Rewrite ($\beta=64$) &-21.10$\pm$17.20&0.85$\pm$0.25&0.76$\pm$0.33 \\ \hline
Rewrite ($\beta=128$) &-22.19$\pm$16.99&0.83$\pm$0.27&0.78$\pm$0.32 \\ \hline
\end{tabular}


\begin{tabular}{l|r@{\hspace{0.75\tabcolsep}}|r@{\hspace{0.75\tabcolsep}}|r@{\hspace{0.75\tabcolsep}}|r@{\hspace{0.75\tabcolsep}}|r@{\hspace{0.75\tabcolsep}}|r@{\hspace{0.75\tabcolsep}}} \hline
 & \multicolumn{6}{c}{Geolocation} \\ \cline{2-7}
 & IR\textsubscript{NA} & IR\textsubscript{Eu} & IR\textsubscript{Africa} &  IR\textsubscript{Asia} & IR\textsubscript{SA} & IR\textsubscript{Remote} \\ \hline
Original & 0.84$\pm$0.32&0.16$\pm$0.33&0.0$\pm$0.0& 0.11$\pm$0.30&0.05$\pm$0.21&0.01$\pm$0.11\\ \hline
Rewrite ($\beta=2$) & 0.82$\pm$0.33&0.17$\pm$0.35&0.0$\pm$0.0& 0.10$\pm$0.30&0.03$\pm$0.17&0.04$\pm$0.20\\ \hline
Rewrite ($\beta=4$) & 0.79$\pm$0.35&0.17$\pm$0.36&0.0$\pm$0.0& 0.12$\pm$0.32&0.04$\pm$0.18&0.04$\pm$0.19\\ \hline
Rewrite ($\beta=8$) & 0.83$\pm$0.31&0.17$\pm$0.34&0.1$\pm$0.08& 0.12$\pm$0.32&0.02$\pm$0.12&0.03$\pm$0.16\\ \hline
Rewrite ($\beta=16$) & 0.81$\pm$0.34&0.16$\pm$0.34&0.0$\pm$0.0& 0.11$\pm$0.31&0.03$\pm$0.18&0.04$\pm$0.19\\ \hline
Rewrite ($\beta=32$) & 0.80$\pm$0.35&0.18$\pm$0.36&0.0$\pm$0.0& 0.11$\pm$0.31&0.03$\pm$0.15&0.04$\pm$0.19\\ \hline
Rewrite ($\beta=64$) & 0.81$\pm$0.33&0.19$\pm$0.37&0.0$\pm$0.0& 0.12$\pm$0.31&0.03$\pm$0.16&0.03$\pm$0.15\\ \hline
Rewrite ($\beta=128$) & 0.81$\pm$0.34&0.16$\pm$0.35&0.0$\pm$0.0& 0.10$\pm$0.29&0.04$\pm$0.18&0.04$\pm$0.20\\ \thickhline
\end{tabular}
\caption{Comparison of generated job description results with scores of original job descriptions for varying values of $\beta$.} 
\label{table:hyperparameter-search-2}
\end{table*}

\section{Differential Impact of Gender Identification on Recommendation Engine}\label{sec:genderimpact}

To ensure that our model successfully generates unbiased job descriptions, we have devised a systematic evaluation approach focused on quantitatively measuring how well the descriptions align with diversity goals. For the evaluation, we first examined original job descriptions to discern which roles were most susceptible to biases. For the candidate recommendation phase, two distinct profiles were constructed for every candidate:

\begin{enumerate}
    \item The gendered profile: This contains a clear statement of gender, articulated as ``I identify as \{gender\}."
    \item The gender-neutral profile: This profile is taken from the original dataset without any explicit gender markers.
\end{enumerate}

\begin{figure*}[tbhp]
    \centering
    \includegraphics[width=6.3in]{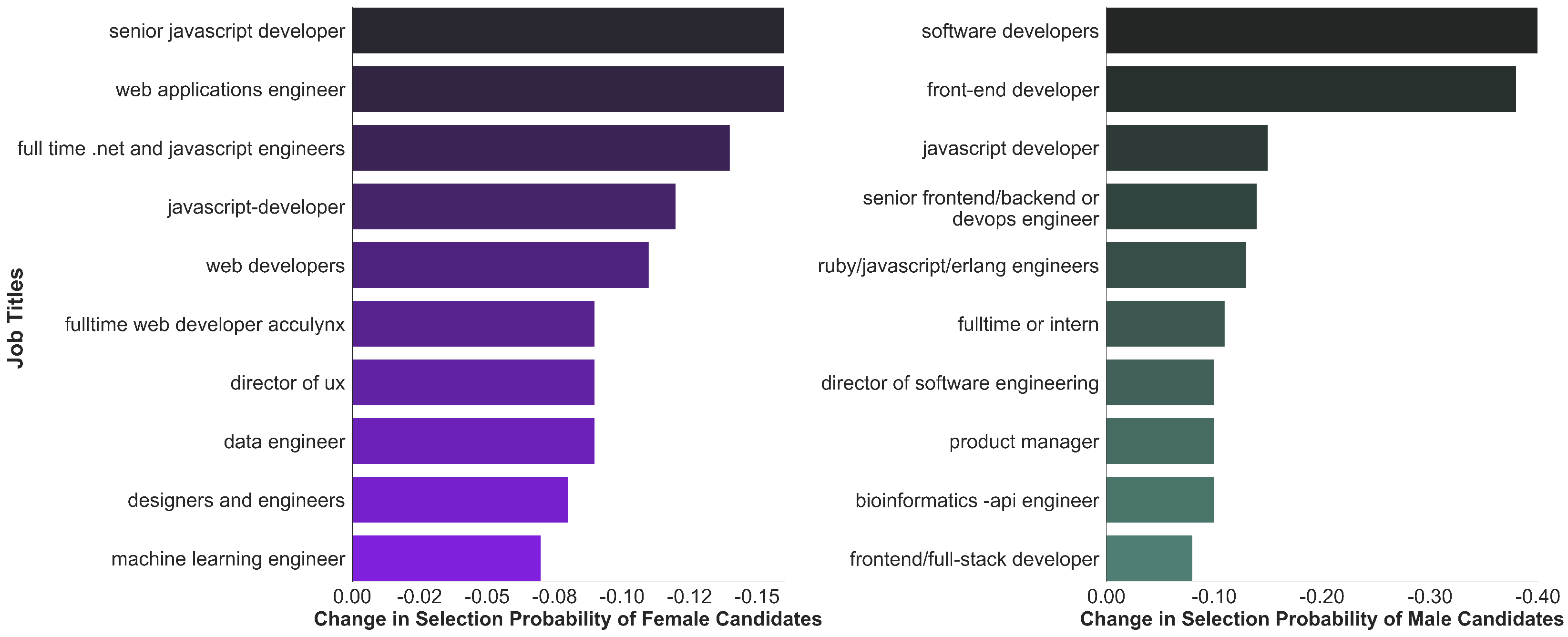}
    \caption{Differential impact on selection probabilities across job titles by gender. This figure visualizes the changes in selection probabilities for various job titles when gender identification is incorporated into candidate profiles. The depicted titles are those experiencing the most pronounced shifts in probabilities. Negative values indicate a reduction in selection probability.}
    \label{fig:genderimpact}
\end{figure*}

Both of these profiles were utilized in the candidate-matching phase with original job descriptions. This approach enabled us to ascertain how gendered or gender-neutral embeddings influenced the recommendation system. We proceeded to determine the disparity in gender distribution (between female and male candidates) matched to each job title utilizing both candidate profile sets of embeddings. The job roles most impacted by gendered embeddings are shared in Figure \ref{fig:genderimpact}.

\section{Introducing Language Quality into Reward Function}

In addition to the reward function presented in Equation 8, we investigated the efficacy of incorporating language quality into the reward function. We adopted the metrics suggested by \citep{zhong2022towards} to assess the coherence, fluency, and relevance of the generated text. Consequently, we define the reward function $\mathcal{R}$ for each generated job description $x$ as:
\begin{equation}
    \mathcal{R}(y_p^*) = \text{LQS}(y_p^*) - \lambda\left(\Delta_\textsubscript{gender}(y_p^*) - \Delta_\textsubscript{geolocation}(y_p^*)\right)
\end{equation}
Here, LQS denotes the language quality metrics, while $\Delta_\textsubscript{gender}$ and $\Delta_\textsubscript{geolocation}$ represent the elements of the diversity score. The parameter $\lambda$ balances language quality with inclusivity considerations. Initially, we fine-tuned the pre-trained GPT-2 for 7 epochs and subsequently trained the perturbation model (RL agent) for 70 epochs. The entire training process spanned 3 days on 8 V100 GPUs. Achieving model convergence with the language score was notably slower than with our primary experimental setup. This delay was attributed to the computational demands of integrating language quality evaluations into the reward function. We showcase the diversity score and impact ratio outcomes for $\beta=64$ and $\lambda=1$ in Table \ref{table:results-language-score-reward}. Our findings indicate that, given the computational costs associated with the two reward functions and the lack of significant improvements in the impact ratio and diversity score, it is more advantageous to solely use the diversity score as a reward function.

\begin{table*}[tbhp]
\centering
\begin{tabular}
{l|r@{\hspace{0.75\tabcolsep}}|r@{\hspace{0.75\tabcolsep}}|r@{\hspace{0.75\tabcolsep}}}
\thickhline
 & \multirow{2}{*}{\shortstack{Diversity \\ score}} & \multicolumn{2}{c}{Gender}  \\ \cline{3-4} 
 & & IR\textsubscript{female} & IR\textsubscript{male} \\ \hline
Original &-23.48$\pm$16.31&0.84$\pm$0.27&0.76$\pm$0.34\\ \hline
Rewrite &-21.97$\pm$16.92&0.83$\pm$0.27&0.81$\pm$0.31 \\ \hline
\end{tabular}
\begin{tabular}{l|r@{\hspace{0.75\tabcolsep}}|r@{\hspace{0.75\tabcolsep}}|r@{\hspace{0.75\tabcolsep}}|r@{\hspace{0.75\tabcolsep}}|r@{\hspace{0.75\tabcolsep}}|r@{\hspace{0.75\tabcolsep}}} \hline
 & \multicolumn{6}{c}{Geolocation} \\ \hline
 & IR\textsubscript{NA} & IR\textsubscript{Eu} & IR\textsubscript{Africa} &  IR\textsubscript{Asia} & IR\textsubscript{SA} & IR\textsubscript{Remote} \\ \cline{2-7} \hline
Original & 0.84$\pm$0.32&0.16$\pm$0.33&0.0$\pm$0.0& 0.11$\pm$0.30&0.05$\pm$0.21&0.01$\pm$0.11\\ \hline
Rewrite & 0.10$\pm$0.29&0.07$\pm$0.25&0.02$\pm$0.12& 0.10$\pm$0.29&0.07$\pm$0.25&0.02$\pm$0.12\\ \thickhline

\end{tabular}
\caption{The results of training with reward function in Equation 1.} 
\label{table:results-language-score-reward}
\end{table*}

\section{Examples}
\label{app:sec:examples}

\begin{table*}[ht]
    \centering
\begin{tabular}{p{10.5cm}|
                r@{\hspace{0.75\tabcolsep}}
                r@{\hspace{0.75\tabcolsep}}r|
                r@{\hspace{0.75\tabcolsep}}
                r@{\hspace{0.75\tabcolsep}}
                r} \thickhline
    \multirow{1}{*}{Description} &
  \multicolumn{3}{c}{Before} &
  \multicolumn{3}{|c}{After} \\ \cline{2-7}
 & $DS$ & $IR_m$ & $IR_f$ & $DS$ & $IR_m$ & $IR_f$ \\ \hline
It'd be a big plus if you have: experience developing games; full health, dental, vision coverage;
-\textcolor{red}{\st{snacked-filled kitchen and booster juice breaks;}} catered breakfast, lunch, and dinner;
- convenient location downtown Toronto
 & -25.35 & 0.44 & 1.00 & -11.35 & 0.25 & 1.00 \\ \hline
XXX, located ... As compensation, we're offering a competitive salary, ..., snacks on snacks \textcolor{red}{\st{ on snacks}}, daily catered lunch, ... & -19.35 & 0.59 & 1.00 & -13.35 & 0.72 & 1.00 \\ \hline
We are hiring exceptional engineers ... are funded by the \textcolor{red}{\st{CEO}} \textcolor{blue}{executive}s of Yelp, Dropbox, Yammer, Box, Parse, and others, as well as Google Ventures, Salesforce and Y-Combinator. Full list at \textcolor{red}{\st{www.}}[REDACTED URL]. Payroll is complex and there are tough engineering challenges to be \textcolor{red}{\st{tack}} \textcolor{blue}{hand}led... 
We strive for 100\% test coverage, and every commit is code review\textcolor{red}{\st{ed by another developer on the team}}... & -19.35 & 0.48 & 1.00 & -13.35 & 1.00 & 0.40 \\ \hline
XXX\textcolor{red}{\st{\textbackslash o/}} - Palo Alto, CA - Full Time ... - H1B OK (visa sort\textcolor{red}{\st{ed}}) XXX captures and indexes every word spoken on TV... and are continuing our m\textcolor{red}{\st{arch}} \textcolor{blue}{ove} onto Google\textcolor{blue}{ }TV and connected devices & -23.35 & 1.00 & 0.61 & -17.35 & 0.85 & 1.00 \\ \hline
... we’re a technology company \textcolor{red}{\st{maniacally focused on}} \textcolor{blue}{dedicated to} a great product. Companies (that you’ve definitely heard of) use Str\textcolor{red}{\st{e}} \textcolor{blue}{ong}ak everyday to make their teams more effective. Future founders, this is a great way to get real experience on what its like starting a company - on our \textcolor{red}{\st{dim}} \textcolor{blue}{not}e...\textcolor{red}{\st{Obvious}} \textcolor{blue}{Unfortunate}ly:... Our benefits package is amazing \textcolor{red}{\st{We are very well funded...}}  & -21.35 & 0.16 & 1.00 & -15.35 & 0.78 & 1.00 \\ \thickhline
\end{tabular}
\caption{Examples of rewrites from the test set showing modifications made by the RL agent along with associated diversity score and impact ratio before and after edits. Relatively minor edits lead to gender inclusivity.}
\label{tb:examples}
\end{table*}

We show some example rewrites along with associated metrics in Tables \ref{tb:examples} and \ref{additional-examples}. In these cases, the rewritten text changes key attributes of the description of the opening (e.g. ``cells and molecules'' to ``databases and documents'', ``Field'' to ``System''), company  (e.g. ``NVIDIA'' to ``MIT'') or stakeholder names (``donor'' to ``investor''), or locations (e.g. ``Princeton'' to ``Paris'') and negations (e.g. ``wont'' to ``would''). This introduces factual errors in the text. For our test sample of $400$ job descriptions, these were relatively infrequent. However, we have not systematically measured hallucinations.

The risk of hallucinations can be mitigated to a large extent as in our intended use case, the rewrite is presented to a recruiter and benefits from human oversight.

\begin{table*}[tbhp]
    \centering
\begin{tabular}{p{10.5cm}|
                r@{\hspace{0.75\tabcolsep}}
                r@{\hspace{0.75\tabcolsep}}r|
                r@{\hspace{0.75\tabcolsep}}
                r@{\hspace{0.75\tabcolsep}}
                r} \hline
    \multirow{1}{*}{Description} &
  \multicolumn{3}{c}{Before} &
  \multicolumn{3}{|c}{After} \\ \cline{2-7}
 & $DS$ & $IR_m$ & $IR_f$ & $DS$ & $IR_m$ & $IR_f$ \\ \hline
Envision a massive, fully-automated research facility that moves around, mixes, and analyzes \textcolor{red}{\st{cells and molecule}} \textcolor{blue}{and databases and documents and papers and thing}s on a scale equivalent to millions of technicians doing the work by hand.  We'll call it the world's first \textcolor{red}{\st{"}}biological server farm\textcolor{red}{\st{"}}--biology will become a programming discipline, and biologists won't need their own labs anymore.
W\textcolor{red}{\st{ant to help us build it?
}} \textcolor{blue}{e're looking for extremely talented software engineers from a variety of backgrounds.  }We're \textcolor{red}{\st{a }}w\textcolor{red}{\st{ell-funded, stealth startup based in Menlo Pa}} \textcolor{blue}{o}rk\textcolor{red}{\st{, founded by scientists and engineers who want to solve biology in their lifetimes.
We're looking for extremely talented software engineers from a variety of backgrounds.  We're working main}} \textcolor{blue}{ing most}ly with C++ and Python in a Linux environment. & -14.85 & 1.00 & 0.37 & -9.85 & 1.00 & 0.62 \\\hline
Santa Clara, CA, Full-time, Linux kernel - Virtualization engineer at \textcolor{red}{\st{NV}} \textcolor{blue}{M}I\textcolor{red}{\st{DIA}} \textcolor{blue}{T}.
We are looking for talented embedded system software engineers with a focus on virtualization to help us architect next generation hypervisor software for \textcolor{red}{\st{NVIDIA platforms}} \textcolor{blue}{the Linux kernel}.
This is a position in Santa Clara, CA.
Some of the skills we look for:
Technical expertise on the ARM architecture, embedded virtualization, \textcolor{red}{\st{multicore}} \textcolor{blue}{divisionot} designs, Linux kernel, device drivers and embedded software in general.
P\textcolor{red}{\st{ractical understanding and implementation of microkernel}} \textcolor{blue}{ practical understanding and implementation of microkot}s, hypervisor design, m\textcolor{red}{\st{ulticore, cache coherency, concurrency, systems level API design, virtual memory management. Also development of virtualization interfaces for the Linux kernel.
Key}} \textcolor{blue}{icroki}wo\textcolor{red}{\st{rds/Specialties: Virtualization, hypervisor design, microkernel}}s, ARM \textcolor{red}{\st{A}} \textcolor{blue}{a}rchitecture, Linux kernel, virtual memory management, Multicore... & -25.35 & 1.00 & 0.76 & -13.35 & 0.62 & 1.00 \\ \hline
GiveNext - C\textcolor{red}{\st{leveland}} \textcolor{blue}{ity}, OH or REMOTE. GiveNext is the easiest way for \textcolor{red}{\st{don}} \textcolor{blue}{invest}ors to give to the causes they care about. We support giving to 1.4 million nonprofits.
Looking for a full-time technical cofounder / CTO. You'll be paid a salary plus have stock options. & -16.35 & 0.92 & 1.00 & -4.85 & 1.00 & 1.00 \\\hline
Daily Harvest - jobs: Software Engineer + more - New York City, NY or P\textcolor{red}{\st{rinceton}} \textcolor{blue}{aris}, NJ | Full-time Onsite | Everyone around you -- especially the non-techies in your life -- will at least try, if not consistently enjoy the \textcolor{red}{\st{frozen superfood eats}} \textcolor{blue}{rocket superfood} that your work at Daily Harvest will deliver! Our 50+ flavor combination\textcolor{red}{\st{s of smoothies, o}} \textcolor{blue}{ of smoothies, instant oops, chia Parfaits, and Har}ve\textcolor{red}{\st{rnight oats, chia parfaits, and har}} \textcolor{blue}{stbowl are co- created by our team of chefs and nutritionists and come packed with organic products and no added added sweet or presews}.... & -20.35 & 1.00 & 0.70 & -9.85 & 1.00 & 0.72 \\\hline
We \textcolor{red}{\st{intend}} \textcolor{blue}{plan} to popularize the production of custom gadgets all over the world. This summer internship is more like an apprenticeship where you learn the \textcolor{red}{\st{rope}} \textcolor{blue}{way}s while following an experienced engineer. ... Monthly stipend \textcolor{blue}{(}R\textcolor{red}{\st{s}}.291\textcolor{red}{\st{6}}7. If you applying from outside India, keep in mind that the total stipend wo\textcolor{red}{\st{nt}}\textcolor{blue}{uld} cover your traveling costs. & -21.35 & 0.37 & 1.00 & -11.35 & 0.43 & 1.00 \\ \hline
*Consulting Engineer (\textcolor{red}{\st{Field}} \textcolor{blue}{System}/implementation/post-sale Engineers) Location: New York, NY / Washington D.C. (Clearance is required)
As a technical consultant, you'll be\textcolor{red}{\st{ MongoDB's ambassador to our clients and other MongoDB users. You'll deliver advisory consulting to and lead comprehensive training sessions with MongoDB's clients, helping them solve mission-critical challenges in areas as varied as schema design, performance optimization (both in a database and in an application), software architecture, production operations. A development/distributed systems background is required.}} & -18.85 & 1.00 & 0.29 & -9.85 & 1.00 & 0.62 \\
\hline
\end{tabular}
\caption{Additional examples with associated metrics.}
\label{additional-examples}
\end{table*}

\end{document}